\documentclass{www2011-accepted}
\pdfoutput=1
\usepackage[usenames,dvipsnames]{color}
\usepackage{amsmath}
\usepackage{xspace,subfigure,url}
\usepackage{xspace}

\newcommand{\xhdr}[1]{\paragraph*{{\bf #1}}}

\newcommand{\certainlong}{certainty\xspace}

\newcommand{\discrep}{Discrepancy}

\newcommand{\filler}{Fillers\xspace}

\newcommand{\ipron}{Indef. pron.}

\newcommand{\tentatlong}{tentativeness\xspace}

\newcommand{\we}{1st pron. pl.}
\newcommand{\welong}{1st person plural pronouns}

\newcommand{\youlong}{2nd person pronoun\xspace}

\newcommand{\Ilong}{1st person singular pronoun\xspace}
\newcommand{\Ilongs}{1st person singular pronouns\xspace}

\newcommand{\positiveemotionsshort}{Pos.\ emo.\xspace}
\newcommand{\positiveemotionslong}{positive emotions\xspace}

\newcommand{\negativeemotionsshort}{Neg.\ emo.\xspace}
\newcommand{\negativeemotionslong}{negative emotions\xspace}

\newcommand{\cat}{C\xspace}
\newcommand{\dimension}{stylistic dimension\xspace}
\newcommand{\dimensionshort}{dimension\xspace}
\newcommand{\dimensions}{style dimensions\xspace}
\newcommand{\dimensionsshort}{dimensions\xspace}

\newcommand{\cohesionfun}
{Coh}
\newcommand{\cohesion}
{stylistic cohesion\xspace}
\newcommand{\Cohesion}
{Stylistic cohesion\xspace}

\newcommand{\accfun}
{Acc}
\newcommand{\User}
{T}
\newcommand{\userpair}
{user pair\xspace}
\newcommand{\userpairs}
{user pairs\xspace}
\newcommand{\T}
{T}
\newcommand{\R}
{R}
\newcommand{\tr}
{t}
\newcommand{\rr}
{r}

\newcommand{\rep}[2]
{#1 \hookrightarrow #2}

\newcommand{\tur}[2]
{#1\leftrightarrow#2}
\newcommand{\influencefun}
{I}
\newcommand{\firstdataset}
{conversational dataset A\xspace}
\newcommand{\Firstdataset}
{Conversational dataset A\xspace}
\newcommand{\seconddataset}
{conversational dataset B\xspace}
\newcommand{\inc}[2]
{#1^{#2}}

\newcommand{\fak}[2]
{#1\not\leftrightarrow#2}
\newcommand{\strictdimensions}
{strictly non-topical \dimensions}
\newcommand{\strictdimension}
{strictly non-topical \dimension}
\newcommand{\numstrictdimensions}
{14\xspace}
\newcommand{\numotherdimensions}
{36\xspace}

\newcommand{\influence}{stylistic influence\xspace}
\newcommand{\Influence}{Stylistic influence\xspace}

\newcommand{\abs}{\mbox{abs}}

\begin{document}
\clubpenalty=10000 
\widowpenalty = 10000

\title{Mark My Words!\\ Linguistic Style Accommodation in Social Media}
\numberofauthors{3}
\author{
\alignauthor 
\hspace*{-.3in}\mbox{Cristian Danescu-Niculescu-Mizil}\titlenote{The research described herein was conducted while
the first author was a summer intern at Microsoft Research.}
       \affaddr{Cornell University}\\
       \affaddr{Ithaca, NY 14853,USA}
       \email{cristian@cs.cornell.edu}
\alignauthor 
Michael Gamon\\
       \affaddr{Microsoft Research}\\
       \affaddr{Redmond, WA 98053,USA}
       \email{mgamon@microsoft.com}
\alignauthor 
Susan Dumais\\
       \affaddr{Microsoft Research}\\
       \affaddr{Redmond, WA 98053,USA}
       \email{sdumais@microsoft.com}
}

\maketitle

\begin{abstract}
The psycholinguistic theory of communication accommodation accounts for the general observation that 
participants in conversations tend to converge to one another's communicative behavior:  they coordinate in a variety of dimensions including choice of words, syntax, utterance length, pitch and gestures.  In its almost forty years of existence, this theory has been empirically supported exclusively through small-scale or controlled laboratory studies.  Here we address this phenomenon in the context of Twitter conversations. Undoubtedly, this setting is unlike any other in which accommodation was observed and, thus, challenging to the theory.  Its novelty comes not only from its size, but also from the non real-time nature of conversations, from the 140 character length restriction, from the wide variety of social relation types, and from a design that was initially not geared towards conversation at all.
Given such constraints, it is not clear \mbox{a priori} whether accommodation is robust enough to occur given the constraints of this new environment.
To investigate this, we develop a probabilistic framework that can model accommodation and measure its effects.  We apply it to a large Twitter conversational dataset specifically developed for this task.
This is the first time the hypothesis of linguistic style accommodation has been examined (and verified) in a large scale, real world setting.

Furthermore, when investigating concepts such as stylistic influence and symmetry of accommodation, we discover a complexity of the phenomenon which was never observed before.  We also explore the potential relation between stylistic influence and network features commonly associated with social status.

\end{abstract}

\vspace{1mm}
\noindent
{\bf Categories and Subject Descriptors:}
J.4 [{\bf Computer Applications}]: {Social and behavioral sciences}

\vspace{1mm}
\noindent
{\bf General Terms:} Measurement, Experimentation, Theory

\vspace{1mm}

\vspace{3in}
\noindent
{\it Language is a social art.}\\ \mbox{\hspace{0.6in}---Willard van Orman Quine}, {\it Word and Object}

\section{Introduction}\label{sec:intro}
The 
theory of communication accommodation was developed to account for the general observation that in conversations people tend to nonconsciously converge to one another's communicative behavior: they coordinate in a variety of dimensions including choice of words, syntax, pausing frequency, pitch and gestures \cite{Giles:1991}.  In the last forty years, this phenomenon has received significant attention and numerous studies indicate that such convergence occurs almost instantly for a very diverse set of communication patterns (see Table \ref{tab:cat} for examples).  
These findings suggest that the communicative behavior of conversational partners are ``patterned and coordinated, like a dance'' \cite{Niederhoffer:2002p2556}.   However, up to now this ``dance'' was exclusively studied in controlled laboratory experiments or through small scale studies.   The work presented here demonstrates for the first time the robustness of accommodation theory in a large scale, real world environment: Twitter.  

\xhdr{Conversations on Twitter: a new hope} Even though not originally developed as a conversation medium, Twitter turns out to be a fertile ground for dyadic interactions.  It is estimated that a quarter of all its users hold conversations with other users on this platform \cite{Java:2007p2731}  and that around 37\% of all tweets are conversational \cite{Ritter:2010p888}.   The fact that these conversations are public renders Twitter one of the largest publicly available resources of naturally occurring conversations.   

Undoubtedly, Twitter conversations are unlike those used in previous studies of accommodation.  One of the main differences is that these conversations are not face-to-face and do not happen in real-time. Like with email, a user does not need to immediately reply to another user's message; this might affect the incentive to use accommodation as a way to increase communication efficiency.  Another difference is the (famous) restriction of 140 characters per message, which might constrain the freedom one user has to accommodate the other.  It is not a priori clear whether accommodation is robust enough to occur under these new constraints.

Also, with very few exceptions, accommodation was only tested in the initial phase of the development of relations between people  (i.e., during the acquaintance process)  \cite{Giles:2008}.  The relations between Twitter users, on the other hand, are expected to cover a much wider spectrum of development, ranging from newly-introduced to old friends (or enemies).  Thus, also from this perspective, the Twitter environment constitutes a new challenge to the theory.

\xhdr{Linguistic style accommodation}  One of the dimensions on which people were shown to accommodate is linguistic style \cite{Niederhoffer:2002p2556,Taylor:2008p3340,Gonzales:2010p3341}, where style denotes the components of language that are unrelated to content:  \textbf{how} things are said as opposed to \textbf{what} is said.  This work will focus on this type of accommodation.  This is a rather important dimension, since, even though only 0.05\% of the English vocabulary is composed of style words (such as articles and prepositions), an estimated 55\% of all words people employ are style words \cite{Tausczik:2010p1495}.  These numbers do not necessarily hold on Twitter, where one might expect style to be sacrificed in favor of content given the length constraint; however, some recent studies also advocate for the importance of style in Twitter  \cite{Eisenstein:2010p2159,Kiciman:2010p2506,Ramage:2010p2455}.   Linguistic style has also been  central to a series of NLP applications like authorship attribution and forensic linguistics \cite{Mosteller:1963p3035,Yule:1939p2733, Holmes:1994p2812,Juola:2008}, gender detection \cite{Koppel:2002p3082,Mukherjee:2010p2452,Herring:2006p2209}  and personality type detection \cite{Argamon:2005p3097}. 

Linguistic style is also known to be, for the most part, generated and processed nonconsciously \cite{Levelt:1982p3608}, and thus a suitable vehicle for studying the phenomenon of accommodation, which itself is assumed to occur nonconsciously.

\xhdr{Probabilistic framework} Previous work on accommodation relied mainly on simple correlation-based measures.  A new framework is necessary in order to correctly model and measure the effects of linguistic style accommodation in a real world, 
uncontrolled
environment.   
The main desirables from such a framework are: 
\begin{itemize}
\item Comparability: the effects of accommodation on different components of style should be comparable.
\item Expressivity:  the framework should be expressive enough to permit the evaluation of particular properties of accommodation (discussed in Section \ref{sec:socling}).
\item Purity: accommodation should not be confounded with other phenomena.
\end{itemize}

The last of these desirables is probably the hardest to achieve and thus deserves some discussion here.  The main challenge is to distinguish accommodation from the effects of homophily: people that converse are likely to employ a similar linguistic style simply because they know each other.   As detailed in Section \ref{sec:framework}, we 
control for this effect by using the temporal aspect specific to accommodation: a person can accommodate to her conversational partner only after receiving her input.  Another type of potential confusion is that between linguistic style accommodation and topic accommodation; this is avoided in this work by a careful selection of the stylistic features following a methodology employed in psycholinguistic literature (discussed in Section \ref{sec:liwc}).  

\xhdr{Stylistic influence and symmetry} Another advantage of the proposed framework is that a new concept of \textit{stylistic influence} emerges naturally: given two conversational partners, one can influence the style of the other more than vice-versa.
This concept is a finer-grained version of the concept of \textit{symmetry} of accommodation proposed in the psycholinguistic literature \cite{Giles:1991}: accommodation can occur symmetrically when both participants in a conversation accommodate to each other or asymmetrically when only one accommodates.  In the latter case, the non-accommodating participant can either maintain her default behavior, or  adjust her behavior in the opposite direction from that of the accommodating participant (i.e., \textit{diverge}).   We are able to show that imbalance in stylistic influence between Twitter users is preponderant and that
symmetry in accommodation is dependent on the stylistic dimension (Figure \ref{fig:symmetry}); for example, users are more likely to accommodate symmetrically on the use of \Ilongs but to accommodate asymmetrically on the use of prepositions.  This is the first time such a rich complexity of the accommodation phenomenon is revealed.

A variety of studies relate accommodation and social status. For example, it was hypothesized  that  a person of lower status will try to accommodate to a person of higher status in order to gain her approval \cite{Giles:2008,Street:1982}.  We take the first steps towards understanding the relation between the concepts of stylistic influence and social status, as reflected in Twitter network features, like number of followers and number of friends, that could be considered (rough) proxies for social status (Section \ref{sec:infexp}).  Rather surprisingly, we observe almost no correlation between these features and stylistic influence.

\xhdr{Applicability}  Apart from its appealing theoretical importance, accommodation also has a variety of potential practical uses.   Based on the premise that accommodation has a subtle positive effect on interpersonal communication, Giles et al. \cite{Giles:2006p2557} discusses applications of accommodation in mediating police-civilian interactions.  On a similar note, Taylor and Thomas \cite{Taylor:2008p3340} shows its relevance in the context of  hostage negotiations. Accommodation was also shown to be practical in the treatment of mental disability \cite{Hamilton:1991} and psychotherapy \cite{Ferrara:1991}.  In Section \ref{sec:conclusions} we also venture into proposing three new potential applications specific to linguistic style accommodation.  We believe that by providing a 
way to model accommodation and by demonstrating its robustness in a real world environment, the present work  provides a 
framework which supports
a wider implementation of such applications.

\section{Communication accommodation theory}\label{sec:socling}
The psycholinguistic theory of communication accommodation was developed around the following main hypothesis:  in dyadic conversations the participants converge to one another's communicative behavior in terms of a wide range of dimensions \cite{Giles:1991}, both verbal and non-verbal.  Table \ref{tab:cat} provides a sample of such converging dimensions.  Many studies seem to indicate that the communicative behaviors of the participants ``are patterned and coordinated, like a dance'' \cite{Niederhoffer:2002p2556}.

\begin{table}
\center
\caption{Examples of dimensions for which accommodation was observed and the respective studies.\nocite{Jaffe:1970}\nocite{Matarazzo:1973}\nocite{Derlega:1973p2564}\nocite{Derlega:1973p2564}\nocite{Hale:1984p2571}\nocite{White:1989p2569}\nocite{Niederhoffer:2002p2556}\nocite{Condon:1967}}
\begin{tabular}{|l|l|}
\hline
\textbf{Dimension} &  \textbf{Canonical study}\\
\hline
Posture	& Condon and Ogston, 1967  \\
Pause length & Jaffe and Feldstein, 1970  \\
Utterance length	& Matarazzo and Wiens, 1973\\
Self-disclosure	& Derlenga et al., 1973  \\
Head nodding	& Hale and Burgoon, 1984\\
Backchannels & White, 1989 \\
Linguistic style &	Niederhoffer and Pennebaker, 2002  \\
\hline
\end{tabular}
\label{tab:cat}
\end{table}

Among various properties of accommodation discussed in the literature, here we briefly review a few that are relevant to our work.  
First, one should keep in mind that the coordination occurs nonconsciously.
Second, accommodation does not necessarily occur \textit{simultaneously} on all dimensions, as shown in \cite{Ferrara:1991}.  Moreover, convergence on some dimensions does not exclude divergence on others: for example, \cite{Bilous:1988} showed that when conversing with males, females converged on frequency of pauses but diverged on laughter.  Another property that is relevant to this work is that of \textit{symmetry} of accommodation: accommodation can occur symmetrically when both participants in a conversation accommodate to each other or asymmetrically when only one accommodates. For example, White \cite{White:1989p2569} presents a study in which Americans accommodate to Japanese on the frequency of backchannels (e.g., `hmm', `uh-huh') but the Japanese did not reciprocate.  Asymmetric accommodation has two flavors, depending on the behavior of the non-accommodating participant: 
\begin{itemize}
\item Default asymmetry: the non-accommodating participant maintains her default behavior (like in the previous example);
\item Divergent asymmetry: the non-accommodating  participant adjusts her behavior in the opposite direction from that of the accommodating participant (i.e., \textit{diverges}) \cite{Giles:1991}.
\end{itemize}  

It is also worth pointing out that the subject of this work is \textit{instant} accommodation, occuring from one conversational turn to another.  Long-term accommodation is considered to be a separate phenomenon with potentially different properties \cite{Ferrara:1991,Giles:1991}. With a few notable exceptions \cite{Ferrara:1991,Niederhoffer:2002p2556}, empirical support for long-term accommodation is absent mostly due to the necessity of longitudinal data.

Various potential explanations for why accommodation occurs have been proposed.  One hypothesis is that accommodation occurs from a desire to increase communicational efficiency \cite{Street:1982}.
Another hypothesis is that a person's convergence to another person's communicative patterns is (nonconsciously) driven by the desire gain the other's social approval \cite{Giles:2008,Street:1982}.  Yet another possible motivation is that accommodation is used to ``maintain a positive social identity'' \cite{Infante:2003} with the other.  The last two hypotheses and several other studies draw a clear relation between social status and accommodation (see also \cite{Giles:1991}), which will become relevant later in our discussion.

In the present work the focus is on \textit{linguistic style accommodation}, and therefore the work of Niederhoffer and Pennebaker \cite{Niederhoffer:2002p2556} is particularly relevant, being the first study to quantify this phenomenon.  It consists of two controlled laboratory experiments (involving 94 dyads) 
and one study based on transcripts of the Watergate tapes (conversations between Nixon and 3 of his aides) in which coordination on various linguistic style dimensions, like usage of prepositions, adverbs and tentative words is shown to occur between the participants.

In its almost forty years of existence, communication accommodation theory was empirically supported exclusively through small scale studies or controlled laboratory experiments.  Also the respective studies focused mainly on real-time interactions (mostly face-to-face, but sometimes computer mediated like in \cite{Niederhoffer:2002p2556}).  With this work we aim to change this state of affairs and demonstrate the robustness of this theory in a large scale, real world environment where conversations are not as richly supported as they are in real-time interactions.

\section{Conversational data}\label{sec:data}
As discussed in Section \ref{sec:intro},  Twitter is a good environment for our study not only because of its fertility in dyadic interactions, but also because it poses new challenges to the theory of communication accommodation in terms of robustness.

Drawing from this resource, Ritter et al. \cite{Ritter:2010p888} builds the largest conversational corpus available to date, made up of 1.3 million conversations between 300,000 users. We will refer to this corpus as \textit{\firstdataset.}  In spite of its size, this corpus presents some major drawbacks with respect to the purpose of this paper.  First, it has a low density of conversations per pair of conversing users:  on average only 4.3 conversations per user; this is not sufficient to model the linguistic style of each pair individually (as required by the accommodation framework proposed in this work and detailed in Section \ref{sec:framework}).   Also, more than half of the pairs of users in this dataset only have unidirectional  interaction, i.e., one of the users in a pair never writes to the other.  This would not introduce a bias with respect to the type of  conversations and relations studied (unidirectional interaction
are generally not classified as normal conversations), but would also drastically limit the potential to compare accommodation between users.

To overcome these limitations, we construct a new conversational dataset with very high density of conversations per pair and with reciprocated interactions.  We start from \firstdataset and select all pairs in which both users initiated a conversation at least 2 times.  We then collect all tweets posted by these users using the Twitter API\footnote{http://apiwiki.twitter.com/} and then reconstruct all the conversations between the selected pair.  The resulting dataset contains 15 million tweets which make up the complete\footnote{Complete up to a maximum 3200 most recent tweets per user, a limitation imposed by the Twitter API.} public twitter activity (a.k.a. public timeline) of 7,800 users; for each user Twitter metadata (such as the number of friends, the number of followers, the location, etc.) is also available.  From these tweets we reconstructed 215,000 conversations between the 2,200 pairs of users with reciprocal relations selected from \firstdataset, using the same methodology for reconstructing conversations employed in \cite{Ritter:2010p888}\footnote{Additionally, we remove self replies and retweets from the data on the belief that they do not make part of a proper dyadic interaction.}.  This conversational dataset is \textit{complete}, in the sense that all twitter conversations ever held within each pair are available.  To the best of our knowledge, this is the largest 
complete conversational dataset.  

The diversity of the user relations and conversations contained in this conversational dataset, dubbed \textit{\seconddataset.}, is illustrated in the following table summarizing  per-pair statistics:
\begin{center}
\small{
\begin{tabular}{|l|c|c|c|c|}
\hline
 & Mean & Median & Min & Max\\
\hline
Number of conversations & 98 & 60 & 1 &1744 \\
Average\ number\ of turns  & 2.7 & 2.6 & 2 & 16.8 \\
Days of contact & 270 & 257 &1 & 886\\
\hline
\end{tabular}}
\end{center}
\normalsize

The main unit of interaction in this work is a \textit{conversational turn}, which is defined as two consecutive tweets in a conversation.  The two tweets in a turn are always sent by different users and are not re-tweets. \Firstdataset contains 2.6 million turns and \seconddataset contains 420,000 turns.\footnote{We are unable to make the data public at the time of publication in consideration of the Twitter terms of service.}

\section{Measuring linguistic style}\label{sec:liwc}

Miller \cite{Miller1996} shows that style and topic are processed differently in the brain. The distinction between the two is important in our investigation of linguistic style accommodation.  In order to measure style and avoid confusion with topic we follow a psycholinguistic methodology used in a variety of applications, known as the LIWC   Linguistic Inquiry Word Count (LIWC) method.

LIWC \cite{liwc}  measures word use in psychologically meaningful categories  (e.g., articles, auxiliary verbs, positive emotions).  It uses over 60 such categories, and dictionaries of words related to each category.
This method has been used in a variety of applications (summarized in \cite{Tausczik:2010p1495}) including to identify social relations, mental health, and individual traits such as gender, age and relative status.  More importantly LIWC is the basis of all recent work on linguistic style accommodation \cite{Niederhoffer:2002p2556,Taylor:2008p3340,Gonzales:2010p3341} to which we want to relate.   

Following the example of these studies, we eliminate all categories related to topic, such as \textit{Leisure}, \textit{Religion} or \textit{Death}.  We refer to the 50 remaining dimensions as \textit{\dimensions.}  
In order to facilitate the presentation of the empirical results, we will focus our discussion on a subset of \numstrictdimensions \dimensionsshort that we call \textit{\strictdimensions:}

\begin{center}
{\small{
\begin{tabular}{|l|l|l|}
\hline
Dimension&Examples&Size\\
\hline
{Article}&an, the&3\\
{Certainty}&always, never&83\\
{Conjunction}&but, whereas&28\\
{Discrepancy}&should, would&76\\
{Exclusive}&without, exclude&17\\
{Inclusive}&with, include&18\\
{Indefinite pronoun}&it, those&46\\
{Negation}&not, never&57\\
{Preposition}&to, with&60\\
{Quantifier}&few, much&89\\
{Tentative}&maybe, perhaps&155\\
{1st person singular pronoun}&I, me&12\\
{1st person plural pronoun}&we, us&12\\
{2nd person pronoun}&you, your&20\\
\hline
\end{tabular}}}\\
\end{center}
\normalsize
For completeness, we mention that all the results presented in this paper also holds for all the other \dimensions (see \cite{Tausczik:2010p1495} for a complete list), unless otherwise noted.  

We say that a tweet \textit{exhibits} a given \dimension  if it contains at least one word from the respective LIWC vocabulary. A tweet can exhibit multiple \dimensionsshort and, in fact, the vast majority do. 

Although we experimented with different methods of extending the LIWC vocabularies with Twitter-specific expressions, we prefer to keep in line with previous literature on linguistic style matching by using the original vocabularies.

\section{Probabilistic framework}\label{sec:framework}
\normalsize{
This section introduces a probabilistic framework that can model the phenomenon of accommodation.  In defining such a framework, the desirable properties discussed in Section \ref{sec:intro} are accounted for: comparability, expressivity and purity.  Although designed to be applicable to any type of conversational data and \dimensions, for notational consistency with the rest of the paper, we use the term ``tweet'' to refer to a conversational utterance.}

\subsection{\Cohesion}\label{sec:cohesion}
\normalsize{
We start by addressing the more general phenomenon of \textit{\cohesion.}  It reflects the intuition that tweets belonging to the same conversation are closer stylistically than tweets that do not.  Cohesion is defined by comparing the probability that a stylistic dimension is exhibited in tweets that are part of a conversation with the probability that the same dimension is exhibited in unrelated tweets. If the former equals the latter, it means that the distribution of the stylistic dimension is the same whether tweets are part of a conversation or not. If the former is larger than the latter, it means that tweets in a conversation tend to ``agree'' with respect to the \dimension.  If the former is smaller than the latter, it means that tweets in a conversation tend to ``disagree'' with respect to the \dimension. Formally, for a given \dimensionshort $\cat$, the measure of \cohesion can be expressed through the following probabilistic expression: }

{\small
\begin{equation}\label{eq:cohesion}
\cohesionfun(\cat)\triangleq {P\left(\T^{\cat}\wedge \R^{\cat}\mid \tur\T\R\right) - P\left(\T^{\cat}\wedge \R^{\cat}\right)}
\end{equation}
}\normalsize where $\T^{\cat}$  (respectively $\R^{\cat}$) is the 
event in which a tweet $\T$ (respectively $\R$) 
exhibits
$\cat$, and $\tur\T\R$ is the condition that tweets $\T$ and $\R$ form a conversational turn\footnote{The sample space considered throughout this work is the set of all possible ordered conversational tweet pairs.}.  Thus, demonstrating that cohesion is observable for \dimension $\cat$ is reduced to showing that $\cohesionfun(\cat)>0$.

It should be emphasized that accommodation is only one of the possible causes for \cohesion.  Another explanation can be the indirect effect of homophily already discussed in Section \ref{sec:intro}: people that converse are likely to employ a similar linguistic style simply because they know each other or are like each other (we will refer to this as \textit{background} style similarity). This observation motivates the need for a measure which can exclusively target accommodation, discussed next.

\subsection{Stylistic accommodation}

When defining a probabilistic framework for linguistic style accommodation it is important to control for the effects of \textit{background} style similarity (and provide the purity desiderata introduced in Section \ref{sec:intro}).  Here this is achieved by measuring accommodation for each \userpair separately and by taking into account the distinctive temporal nature of accommodation: a user can accommodate to her conversational partner only after receiving her input.
In doing so, the concern is eliminated because a confusion with background style similarity effects, like homophily, would not be expected to cause differences within a single pair depending on whether 
one or the other user in a pair
initiates a conversational turn.
Therefore, the goal is to measure for a given pair of users $a$ and $b$ who engage in a conversation whether the use of a \dimension $\cat$ in the initial tweet (of user $a$) increases the probability of that stylistic dimension in the reply (of user $b$) beyond what is normally expected from user $b$ (when replying to user $a$).

Formally, for a given \dimension $\cat$ and pair of users $(a,b)$, the  accommodation  of user $b$ to user $a$  is measured by how much the fact that user $a$ exhibits $\cat$ in a tweet $\T_a$ increases the probability of $b$ to also exhibit $\cat$ in a reply to $\T_a$:
\vspace{-0.1in}

\small
\begin{equation}
\accfun_{(a,b)}(\cat)\triangleq {P\left(\User_b^{\cat} \mid \User_a^{\cat},  \rep{\User_b}{\User_a}\right)-P\left(\User_b^{\cat} \mid \rep {\User_b}{\User_a}\right)}\label{eq:accdef}
\end{equation}
\normalsize
where $\User_a^{\cat}$ (respectively $\User_b^{\cat}$) is the
event in which
a tweet posted by user $a$ (respectively $b$)  exhibits $\cat$, and $\rep{\User_b}{\User_a}$ is the condition that $\T_b$ is a reply to $\T_a$.  This condition, present in both the minuend and the subtrahend\footnote{Where by minuend we mean the left term of a subtraction and by subtrahend we mean to the right one.}, has the role of restricting this measure of accommodation only to replies of $b$ to  $a$, therefore controlling for differences in the background linguistic similarity between users.  Also note that by using the $\rep{}{}$ condition instead of the $\tur{}{}$ condition employed in the definition of cohesion (\ref{eq:cohesion}), we embed the distinctive temporal aspect of accommodation mentioned earlier.  This ability to integrate 
temporal disparity
is an essential 
advantage 
of this framework over the correlation based measures previously used in studies of stylistic accommodation \cite{Niederhoffer:2002p2556,Taylor:2008p3340,Gonzales:2010p3341}\footnote{As a concrete example,  correlation does not distinguish between the the case in which the initial tweet exhibits $\cat$ but the reply does not, and the reverse case in which the the initial tweet does not exhibit $\cat$ but the reply does.}.

Since the main goal is to address global accommodation (as opposed to the  within-pair accommodation described above), the accommodation for a given \dimensionshort $\cat$ is defined as:
\begin{equation}
\accfun(\cat)=E[\accfun_{(a,b)}(\cat)]
\end{equation} 
where the expectation is taken over all possible conversing pairs $(a,b)$.
Under this framework, proving that accommodation is observable for \dimension $\cat$ is reduced 
to showing that $\accfun(\cat)>0$.

\subsection{Stylistic influence and symmetry}\label{sec:influence}
\normalsize
One important property of the way $\accfun$ is defined is its asymmetry: the accommodation of user $b$ to user $a$ on \dimension $\cat$  is potentially different from the accommodation of user $a$ to user $b$ on the same \dimension.  The notion of stylistic influence arises naturally:
\begin{equation}\label{eq:influencefun}
\influencefun_{(a,b)}(\cat)\triangleq \accfun_{(a,b)}(\cat) - \accfun_{(b,a)}(\cat)
\end{equation}
for a given \dimension \cat.  
If $\influencefun_{(a,b)}(C)>0$ we can say that $b$ accommodates more to $a$ on $\cat$ than $b$ does to $a$.  

A related concept is
accommodation symmetry (discussed in Section \ref{sec:socling}), which 
is tied to
to the 
accommodation measure in the following way.  Given that $b$ accommodates to $a$, i.e $\accfun_{(a,b)}(\cat)>0$, we have
\begin{itemize}
\item Symmetry when {$\accfun_{(b,a)}(\cat)>0$},
\item Default asymmetry when {$\accfun_{(b,a)}(\cat)=0$},
\item Divergent asymmetry when: {$\accfun_{(b,a)}(\cat)<0$}
\end{itemize}

\section{Empirical validation}
Equipped with the probabilistic framework introduced in the previous section, here we proceed with an empirical validation of the accommodation phenomenon on the conversation data described in Section \ref{sec:data}.  As previously discussed, this setting is fundamentally different from all other circumstances in which the theory of communication accommodation was validated, therefore challenging its robustness.

\subsection{Validation of \cohesion} 
We start by asking whether Twitter conversations are characterized by \cohesion, since this is a precondition for accommodation.  The \cohesion model described in Section \ref{sec:cohesion} does not distinguish between users and therefore can be directly applied to the \firstdataset(introduced in Section \ref{sec:data}).

In order to demonstrate that cohesion is exhibited in our data 
we estimate
the two probabilities involved in  (\ref{eq:cohesion}) as follows.
We estimate the first probability as
the fraction of all turns in which both tweets exhibit \dimensionshort $\cat$: 

{\small
\begin{equation}
\widehat{P}\left(\T^{\cat}\wedge \R^{\cat}\mid \tur \T \R\right)=\frac{|\left\{(\tr,\rr)\mid\tur \tr \rr,\inc\tr\cat,\inc\rr\cat\right\}|}{|\left\{(\tr,\rr)\mid \tur \tr \rr\right\}|}
\end{equation}}
where $\inc\tr\cat$ denotes the condition that a tweet $t$ exhibits  \cat.\footnote{Lowercase letters are used to represent tweets that make up our dataset, distinguish them from the uppercase letters that refer to probabilistic events in the framework defined in Section \ref{sec:framework}.}

To estimate the second probability, we first construct a set of ``fake turns'' by randomly pairing together tweets from the entire conversational data (regardless of their authors).  We can then write:
{\small
\begin{equation}
\widehat{P}\left(\T^{\cat}\wedge \R^{\cat}\right)=\frac{|\left\{(\tr,\rr)\mid\fak \tr \rr,\inc\tr\cat,\inc\rr\cat\right\}|}{|\left\{(\tr,\rr)\mid \fak \tr \rr\right\}|}
\end{equation}
}where $\inc\tr\cat$ is the condition that the tweet $\tr$ exhibits $\cat$ and  $\fak \tr \rr$ is the condition that the tweets $\tr$ and $\rr$ are paired together in a fake turn.

Establishing that cohesion is exhibited in the data 
corresponds to rejecting the null hypothesis of these two probabilities being equal.  
Fisher's exact test\footnote{We use this exact variant of the $\chi^2$ test since for some \dimensions the expected counts are low.}
rejects this hypothesis with p-value smaller than $0.0001$ for each of the \strictdimensions.
 
\begin{figure}[t!]
\includegraphics[width=3.3in]{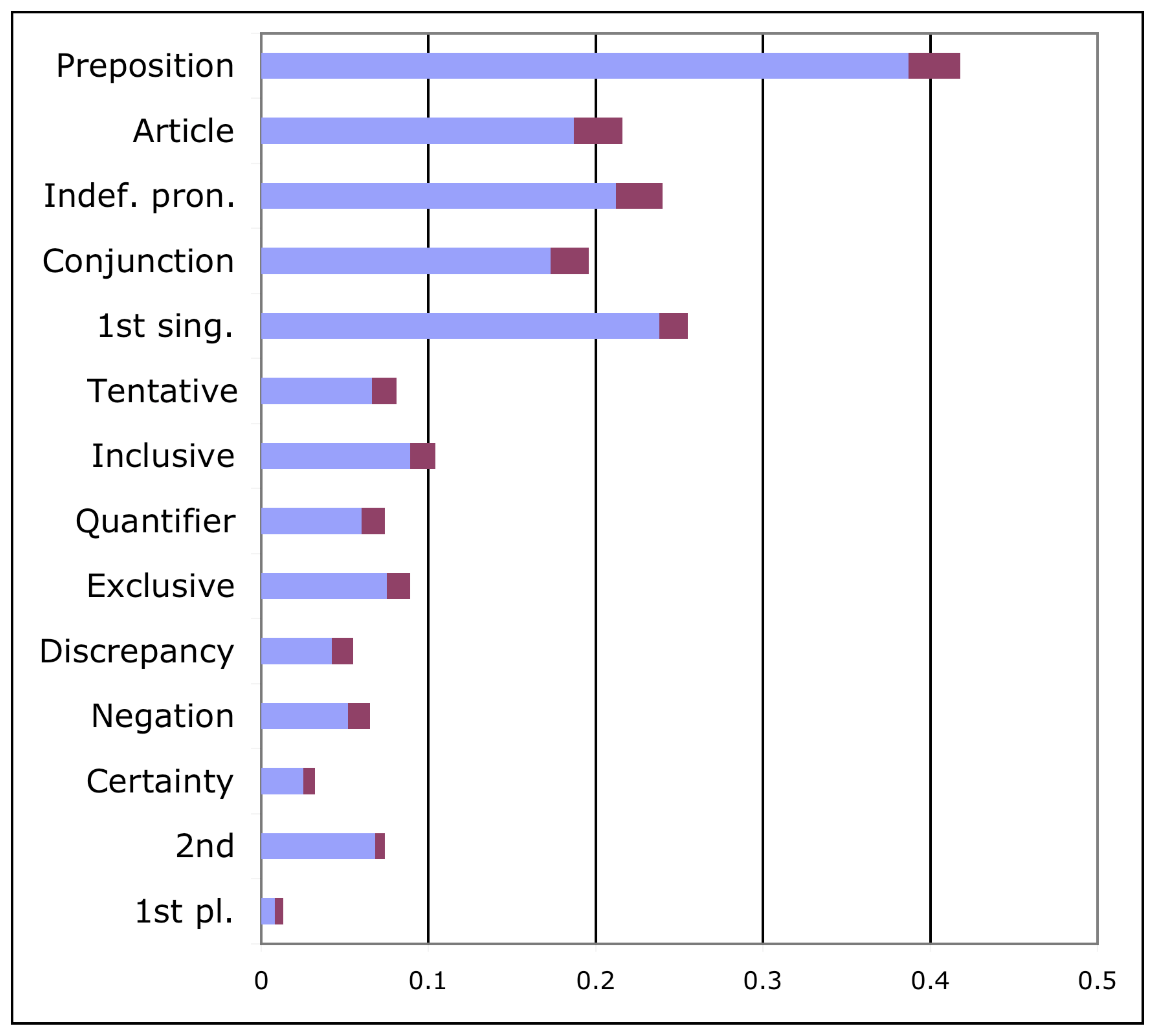}
\caption{The effect of \cohesion observed as the difference between $\widehat{P}\left(\T^{\cat}\wedge \R^{\cat}\mid \tur \T \R\right)$ (composite bars) and $\widehat{P}\left(\T^{\cat}\wedge \R^{\cat}\right)$ (blue bars). The differences, shown in red/dark, are statistically significant (p<0.0001). The \dimensionsshort are shown in decreasing order of the difference.}

\label{fig:cohesion}
\end{figure}

Figure \ref{fig:cohesion} shows the 
estimates of the two probabilities for each of these \dimensions 
(the difference between the two  is shown in red/dark).  
While this result is not surprising, it is a necessary precondition for verifying the more subtle hypothesis of accommodation that we are going to address next.

\subsection{Validation of stylistic accommodation}\label{sec:accommodation}

We now proceed to answer the main question of this work: does the hypothesis of stylistic accommodation proposed in the psycholinguistic literature hold in social media conversations?  Since the probabilistic framework for accommodation is applied at the level of user pairs, the \seconddataset is employed for this analysis.

For each ordered \userpair $(a,b)$ and \dimension $\cat$, we estimate the minuend in (\ref{eq:accdef}) as the fraction of $b$'s replies to $a$ in which $b$'s tweet  $\tr_b$ exhibits $\cat$:

\small{
\begin{equation}
\widehat{P}\left(\User_b^{\cat} \mid \rep {\User_b} {\User_a}\right)=\frac{|\left\{(\tr_a,\tr_b)\mid \rep {\tr_b} {\tr_a},\inc{\tr_b}\cat\right\}|}{|\left\{(\tr_a,\tr_b)\mid \rep {\tr_b} {\tr_a}\right\}|}
\end{equation}
}
\normalsize
Similarly, the subtrahend is estimated as:
\small{
\begin{equation}
\widehat{P}\left(\User_b^{\cat} \mid \User_a^{\cat},  \rep {\User_b} {\User_a}\right)=\frac{|\left\{(\tr_a,\tr_b)\mid \rep {\tr_b} {\tr_a}, \inc{\tr_b}\cat,\inc{\tr_a}\cat\right\}|}{|\left\{(\tr_a,\tr_b)\mid \rep {\tr_b} {\tr_a}, \inc{\tr_a}\cat\right\}|}
\end{equation}
}
\normalsize

\begin{figure*}[th]
\includegraphics[width=6.8in]{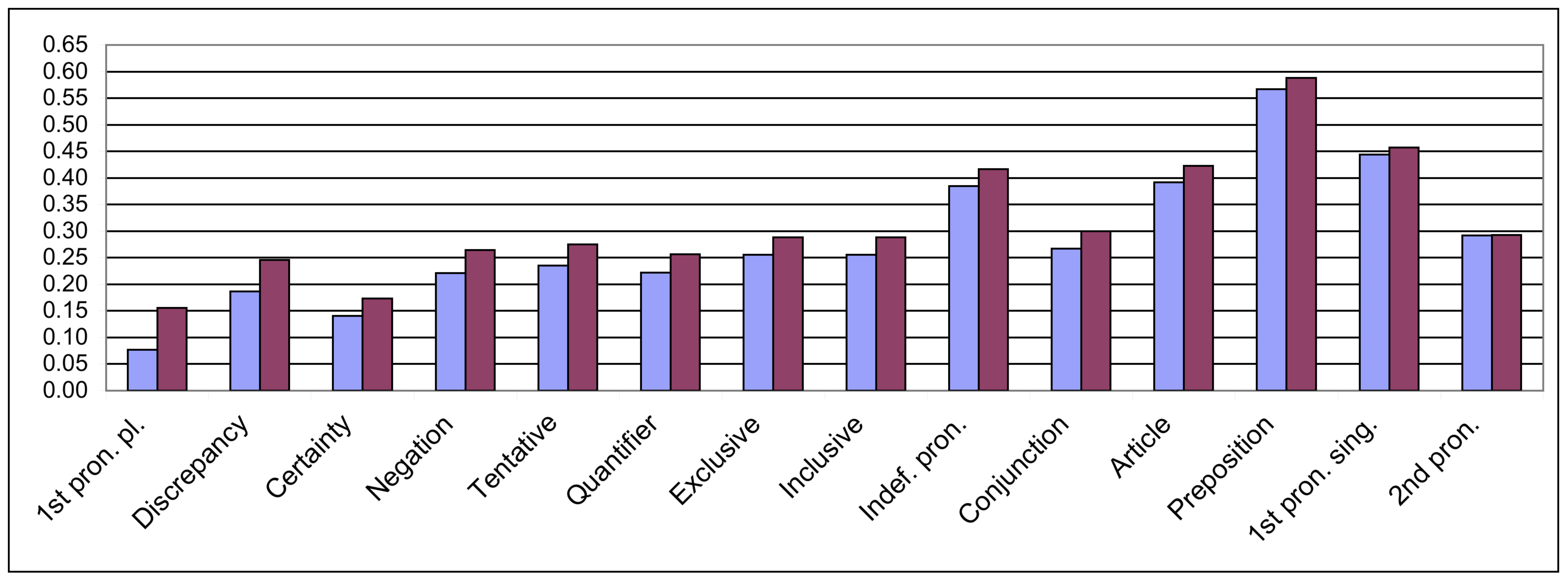}
\caption{The effect of accommodation $\widehat\accfun(\cat)$ for each \strictdimension $\cat$ observed as the difference between the means of $\left\{\widehat{P}\left(\User_b^{\cat} \mid \rep{\User_b},{\User_a}\right)\mid (a,b)\in \mbox{Pairs} \right\}$ (blue, left) and $\left\{\widehat{P}\left(\User_b^{\cat} \mid \User_a^{\cat},  \rep{\User_b}{\User_a}\right)\mid (a,b)\in \mbox{Pairs} \right\}$ (red, right).  All the differences are statistically significant (p<0.0001), except for the \textit{\youlong} category. The \dimensionsshort are ordered according to the amount of accommodation observed.}\label{fig:accommodation}
\end{figure*}

We can then measure the amount of accommodation $\widehat{\accfun}(\cat)$ exhibited in our dataset as the difference between the mean of the set of subtrahend estimations 
$$\left\{\widehat{P}\left(\User_b^{\cat} \mid \User_a^{\cat},  \rep{ \User_b} {\User_a} \right)\mid  (a,b) \in \mbox{Pairs}\right\}$$ 
 and the mean of the minuend estimations$$\left\{\widehat{P}\left(\User_b^{\cat} \mid \rep{\User_b}{ \User_a} \right)\mid  (a,b) \in \mbox{Pairs}\right\},$$ where $\mbox{Pairs}$ is the set of all ordered pairs\footnote{We discard all \userpairs for which the denominator of any of these two estimations is less than 10.}.
Figure \ref{fig:accommodation} compares these means 
--- the former is illustrated in red/right, the latter in blue/left --- 
for each \strictdimension.  All the differences 
are statistically significant with a p-value smaller than 0.0001 according to a two-tailed paired t-test\footnote{In order to allay concerns regarding the independence assumption of this test, for each two users $a$ and $b$ we only consider one of the two possible ordered pairs  $(a,b)$ and $(b,a)$.} 
for all \strictdimensions with the exception of the \textit{\youlong} \dimension for which the difference is not statistically significant.

Even though our focus is on the \strictdimensions, for completeness we also measured accommodation on the remaining \numotherdimensions \dimensionsshort and observed a statistically significant effect for all of them except for \textit\filler (like `blah',\ `yaknow') for which the data was insufficient.

Note that by design, our probabilistic framework allows comparison between the accommodation effects exhibited for each dimension $\cat$ (i.e., fulfills the comparability desiderata introduced in Section \ref{sec:intro}).  Here are some of the comparisons worth pointing out:

\begin{itemize}
\item Users accommodate significantly more on \textit\tentatlong than on \textit\certainlong  (p-value smaller than $0.01$ according to an independent t-test).\footnote{Therefore doubt appears to be more ``contagious'' than confidence.}
\item 
Users accommodate significantly more on \textit{\negativeemotionslong}  than on \textit\positiveemotionslong (not illustrated,\\ 
$\widehat{\accfun}(\negativeemotionsshort)=0.07$,\ $\widehat{\accfun}(\positiveemotionsshort)=0.04$;\\
 p-value smaller than $0.01$ according to an independent t-test for the difference).
\item \textit\Ilong vs. \textit{\youlong.}  In retrospect, the fact that accommodation is not exhibited for the \textit{\youlong} \dimensionshort seems natural: words like `you' have a different meaning for two participants involved in a conversation.  However, the same holds for the \textit\Ilong \dimensionshort for which accommodation is observed.  This could be explained by the social-psychology hypothesis of disclosure reciprocity in dyadic relationships \cite{Derlega:1973p2564}.
\end{itemize}

With the results presented here we are able to verify that accommodation does indeed hold in large scale, real world conversational setting with properties that a priori seemed challenging to the theory.  In the remainder of this section we will use our framework to investigate what properties linguistic style accommodation exhibits in this conversational setting.

\subsection{\Influence and symmetry}\label{sec:infexp}
Here we seek to understand the role that the concept of \influence (introduced in Section \ref{sec:influence}) has in Twitter conversations.  We start by asking whether \influence is prevalent in the data: in general, is there a balance between the amount two participants in a conversation accommodate? Or, on the contrary, is one user stylistically dominating the other?

In terms of our framework, we can test whether in expectation there is an imbalance of accommodation between participants in a conversation by verifying whether we can reject the null hypothesis 
$E\left[\abs(\influencefun_{(a,b)}(\cat))\right] = 0$, where the expectation is taken over all conversing pairs $(a,b)$.  Using definition (\ref{eq:influencefun}), this is reduced to rejecting: 
$$E\left[\abs\left(\accfun_{(a,b)}(\cat)-\accfun_{(b,a)}(\cat)\right)\right]=0.$$
and further to rejecting:
\begin{align}
&E\left[\max\left(\accfun_{(a,b)}(\cat),\accfun_{(b,a)}(\cat)\right)\right]=\nonumber\\
&E\left[\min\left(\accfun_{(a,b)}(\cat),\accfun_{(b,a)}(\cat)\right)\right]\nonumber
\end{align}
where the first term is the expected accommodation of the most accommodating users (where the accommodation is always compared within each pair), and can by estimated the mean of:
$$\left\{\max\left(\widehat\accfun_{(a,b)}(\cat),\widehat\accfun_{(b,a)}(\cat)\right)\mid (a,b)\in \mbox{Pairs}\right\},$$ and the second term is the expected accommodation of the least accommodating users, estimated by the mean of:
$$\left\{\min\left(\widehat\accfun_{(a,b)}(\cat),\widehat\accfun_{(b,a)}(\cat)\right)\mid (a,b)\in \mbox{Pairs}\right\}.$$

 \begin{figure*}[ht!]
\includegraphics[width=6.8in]{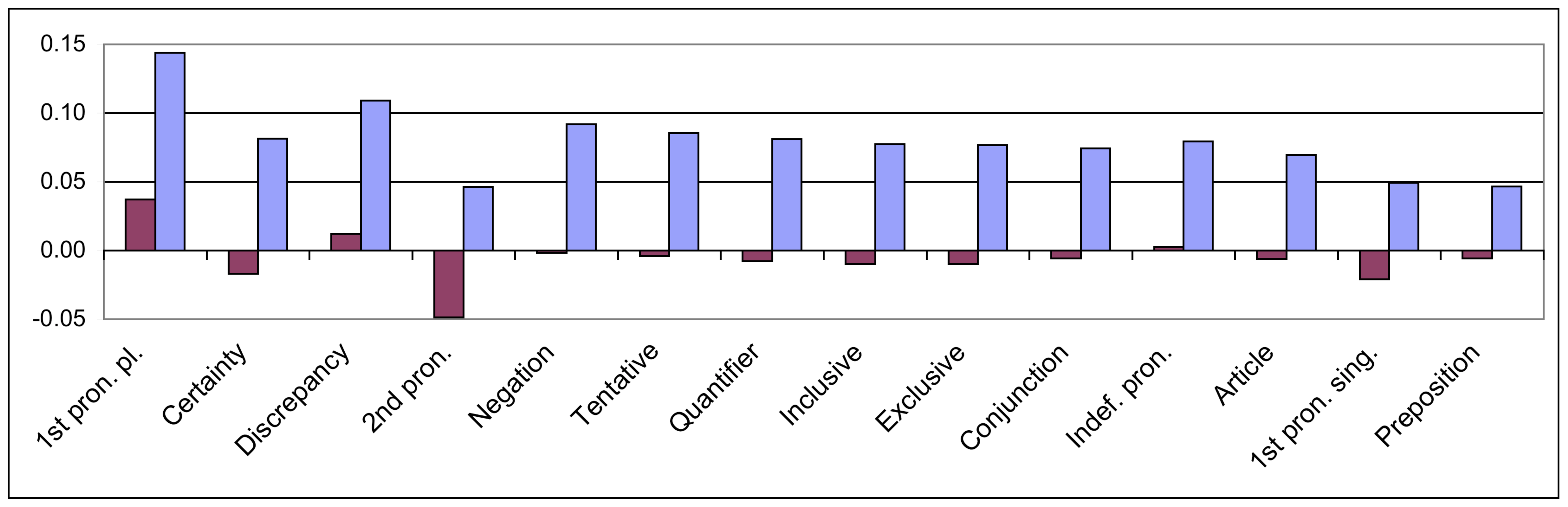}
\caption{The effect of \influence for each \strictdimension $\cat$ observed as the difference between the means of $\left\{\min\left(\widehat\accfun_{(a,b)}(\cat),\widehat\accfun_{(b,a)}(\cat)\right)\mid (a,b)\in \mbox{Pairs} \right\}$ (red, left) and $\left\{\max\left(\widehat\accfun_{(a,b)}(\cat) \right)\mid (a,b)\in \mbox{Pairs} \right\}$ (blue, right).  All the differences are statistically significant (p<0.0001).
The \dimensionsshort are shown in decreasing order of the difference.}\label{fig:influence}
\end{figure*}

Using the same method for estimating $\accfun_{(a,b)}(\cat)$ discussed in Section \ref{sec:accommodation}, we reject this hypothesis for all \strictdimensions $\cat$ (paired t-test with p-value smaller than 0.0001)\footnote{The same holds for all the other dimensions except \textit\filler for which the data was insufficient.}.  Figure \ref{fig:influence} 
illustrates the difference between the expected accommodation of the least accommodating users (red/left) and that of the most accommodating users (blue/right) in a pair. A difference in the 
type
of imbalance between \dimensionshort is revealed;  for example, while for \textit{\welong} in general the least accommodating users 
still match the style of the most accommodating participants (even though significantly less than vice-versa), for \textit{\certainlong} the least accommodating users in general diverge from the style of the most accommodating participants.

\begin{figure}[th!]
\includegraphics[width=3.3in]{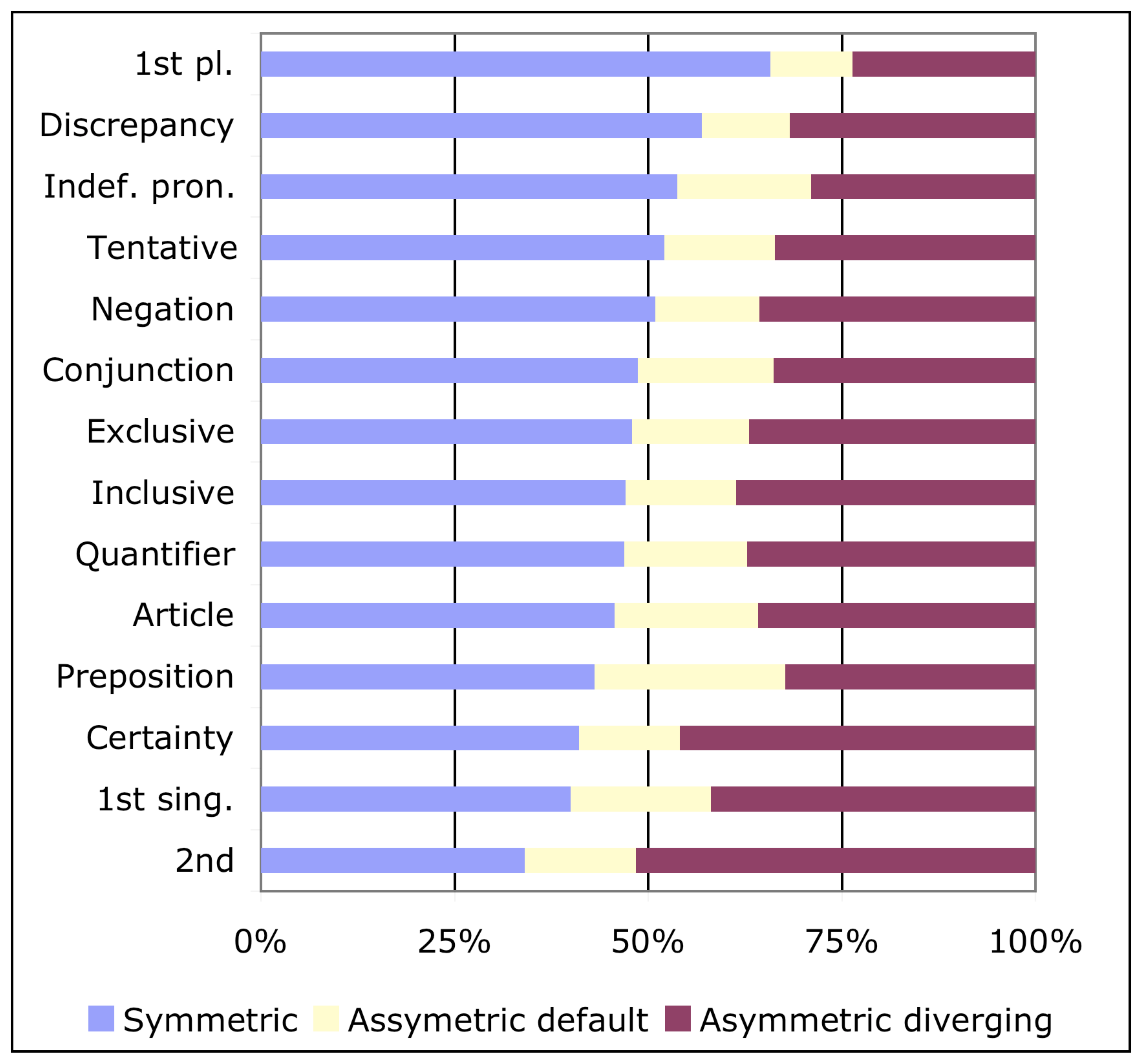}
\caption{The percentage of accommodating pairs that exhibit each of the three types of accommodation: symmetric, default asymmetry and diverging assymetry.}
\label{fig:symmetry}
\end{figure}

To further investigate this intriguing difference between \dimensions, we turn our attention to the property of symmetry.  Figure \ref{fig:symmetry} shows the percentage split between symmetrically accommodating pairs (blue/left), asymmetrically default accommodating pairs (yellow/center) and asymmetrically diverging accommodating pairs (red/right), as defined in Section \ref{sec:influence}.  

The conclusions that can be drawn from analyzing these results is that accommodation is a much more complex behavior than previously reported in the literature, where it was assumed that only one type of accommodation occurs for a given dimension\footnote{Here we refer to any dimension of accommodation, like the ones in Table \ref{tab:cat}, not only to linguistic style dimensions.}.  But as it can be observed in Figure \ref{fig:symmetry} all three types of accommodation have a considerable stake.
Furthermore, in all previous work on linguistic style accommodation, no distinction was made between the type of accommodation occurring for each 
\dimensionshort.  However, our study indicates a clear difference between \dimensionsshort:

\begin{itemize}
\item Symmetric accommodation is dominant for \textit{\we}, \textit{\discrep} and \textit{\ipron};
\item Asymmetric accommodation (of both types) is dominant in most of the other dimensions;
\item Asymmetric diverging accommodation is dominant for \textit{\youlong.}
\end{itemize}

A potential explanation for the fact that such a complex accommodation behavior 
was not previously observed may be the difference between the Twitter conversational setting and that traditionally used in the literature (discussed in Section \ref{sec:intro}), especially in the spectrum of relation types covered (mostly limited to one type in the previous studies).   Another explanation may be the increased expressibility of our probabilistic framework over the correlation based framework used in previous studies.

\subsection{Relation to social status}
As pointed out in Section \ref{sec:socling}, the psycholinguistic literature draws clear a connection between the social status of a user and its  tendency to accommodate.  Therefore, it is natural to ask whether \influence correlates with differences in social status between the users and we take the first steps to address this question.  For lack of a better proxy, we employ user features that were previously reported to be related to social influence on Twitter \cite{Bakshy:2011p4940}.   For each pair of users in our data we compare: \#followers, \#followees, \#posts, \#days on Twitter, \#posts per day and ownership of a personal website.  We find that for all \dimensions none of these features correlate strongly with \influence; the largest positive Pearson correlation coefficient obtained was 0.15 between \#followees and \influence on \textit{\we}.  Also, for the task of predicting the most influential user in each pair a decision tree classifier\footnote{We used the Weka implementation of the C4.5 decision tree, available at \url{www.cs.waikato.ac.nz/ml/weka/}} rendered relatively poor results. The best improvement over  the majority class baseline was of only 7\% for the \textit{\we} \dimensionshort (in this case the most predictive features were the difference in \#friends and the difference in ownership of a personal website).  All this suggests that
 \influence
 
appears to be only weakly connected to these 
social features.  
However, one should take this observation with a grain of salt: the proxies for social status available on Twitter and employed here are far from ideal. Future work should seek to use better proxies for social status, possibly in environments with richer social data.

\section{Related work}

Here we briefly touch on related work not already discussed. Much of the research in understanding social media focuses on the network relations between users.  More recently, this line of work has been complimented with a rich analysis of the content of posts as well as structural relations among posters.   In one early study combing these two dimensions of analysis, Paolillo \cite{Paolillo:2001p2215} examined linguistic variations associated with strong and weak ties in an early internet chat relay system.  The strength of friendship ties on Facebook was related by Gilbert and Karahalios \cite{Gilbert:2009p2560} to various language features including intimacy words, positive and negative emotions.
Eisenstein et al. \cite{Eisenstein:2010p2159} investigated the role geographic variation of language has in Twitter
and Kiciman \cite{Kiciman:2010p2506} examined the extent to which differences in language models of Twitter posts (as measured by perplexity) were related to metadata associated with the senders. 
Such work demonstrates the importance of linguistic style variations in Twitter which also plays a crucial role in our study.
 
Latent variable models have also been used to summarize more general linguistic patterns in social media.  Ramage et al. \cite{Ramage:2010p2455} developed a partially supervised learning model (Labeled LDA) to summarize key linguistic trends in a large collection of millions of Twitter posts.  They identified four general types of dimensions, which they characterized as substance, status, social and style.   These included dimensions about events, ideas, things, or people (substance), related to some socially communicative end (social), related to personal updates (status), and indicative of broader trends of language use (style).    This representation was used to improve filtering of tweets and recommendations of people to follow.    In the task of tweet ranking, a different approach is taken by \cite{Duan:2010p2451} which employs twitter specific features in conjunction to textual content.   Another way of characterizing key trends in text data is to use known distinctions or dimensions. In addition to the already discussed work based on LIWC, see  \cite{Milroy:1992p2730, Paolillo:2001p2215} for other examples of analyses of linguistic variation with respect to position or status in a social network.  Of particular interest in our work is the distinction between linguistic style and content.   Style or function words make up about 55\% of the words that we speak, read and hear according to Tausczik and Pennebaker \cite{Tausczik:2010p1495}, similar to findings of Ramage et al. \cite{Ramage:2010p2455}  in their analysis of Twitter.   In our research, we use LIWC to characterize the linguistic style of posts as well as individuals.

One particularly interesting type of linguistic activity in social media has to do with conversations, that is with exchanges between one or more individuals.  Twitter conversations are the main focus in this work. Java et al. \cite{Java:2007p2731} found that 21\% of users in their study used Twitter for conversational purposes (as measured by the use of @, a convention to address a post to a particular user), and that 12.5\% of all posts were part of conversations. Honeycutt and Herring \cite{Honeycutt:2009p192} analyzed conversational exchanges on the Twitter public timeline, focusing on the function of the @ sign.  They found that short dyadic conversations occur frequently, along with some longer multi-participant conversations.   Ritter et al. \cite{Ritter:2010p888} developed an unsupervised learning approach to identify conversational structure from open-topic conversations.  Specifically they trained an LDA model which combined conversational (speech acts) and content topics on a corpus of 1.3 million Twitter conversations, and discovered interpretable speech acts (reference broadcast, status, question, reaction, comment, etc.) by clustering utterances with similar conversational roles.   In our research, we build on this data set and extend it to include the complete conversational history of individuals over a period of almost one year.

Since the notion of linguistic style is central to this work, we also want to point out other instances in which it plays an important role.  Linguistic style was shown to be crucial in the area of authorship attribution and in forensic linguistics (for an overview see \cite{Juola:2008}). To identify an author, it is necessary to look beyond content into the --- often subconscious --- stylistic properties of the language. Simple measures like word length, word complexity, sentence length and  vocabulary complexity were at the forefront of earlier research into attribution problems (e.g. \cite{Yule:1939p2733,Holmes:1994p2812}). Since Mosteller and Wallace's seminal work on the Federalist Papers \cite{Mosteller:1963p3035}, however, a trend has emerged to focus on the distribution of function words as a diagnostic for authorship, a method that in various incarnations now dominates the research. Other areas using similar methods include gender detection from text \cite{Koppel:2002p3082,Mukherjee:2010p2452,Herring:2006p2209} and personality type detection \cite{Argamon:2005p3097}.

\section{Conclusions and future work}\label{sec:conclusions}
In this paper we have shown that the hypothesis of linguistic style accommodation can be confirmed in a real life, large scale dataset of Twitter conversations.   We have developed a probabilistic framework that allows us to measure accommodation and, importantly, to distinguish effects of style accommodation from those of homophily and topic-accommodation. We also have demonstrated how this framework allows us to formalize and investigate the notions of stylistic influence and asymmetric accommodation. 

It is important to 
highlight
that our findings are anything but obvious,
given that Twitter is a medium 
unlike any other setting in which the phenomenon was previously observed.  Its novelty comes not only from its size, but also from the wide variety of social relation types, from the non real-time nature of conversations, from the 140 character length restriction  and from a design that was initially not geared towards conversation at all.
This work demonstrates that accommodation is robust enough to occur under these new constraints, presumably because it is deeply ingrained in human social behavior.

We believe that this line of research has a number of natural extensions. One question we have not addressed is the issue of long-term accommodation: can we measure accommodation over a longer period of time, from the first interaction of two users on? Answering this question is challenging because it requires richer longitudinal data.  It would also be very interesting to explore interplay between the accommodating behavior and the type of social relation. 

As for practical applications, on the premise that accommodation renders conversations more pleasant and effective, we posit that having the linguistic style of automated dialogue systems match that of the user would increase the quality of the interaction.  Personalized ranking of tweets could also benefit by selecting tweets with styles that match that of the tweets issued by the target user.  Finally, given the evidence that this work brings to support the universality of the accommodation phenomenon, we envision its use in detection of forged conversations.\footnote{We are inspired here by the use of Benford's law in detecting forged financial reports \cite{TamCho:2007p3501}.  Though potentially not as common as such forgery, situations in which conversational transcripts are contested are not infrequent.  One recent example is the October 2010 release of phone conversations between top Romanian political leaders and a compromised business man.  Another one is the controversy surrounding the reality TV shows ``The Jersey Shore'' and ``The Hill''  which are suspected of being scripted.   
}

Finally, we 
hope that our findings will stimulate further research and refinements of the communication accommodation theory in the psycholinguistic world.

\vspace{0.2in}
\noindent{\small
\textbf{Acknowledgments}  We thank Lillian Lee for inspiring conversations, Munmun De Choudhury, Scott Counts, Sumit Basu, Dan Liebling, Magdalena Naro\.zniak, Alexandru Niculescu-Mizil, Tim Paek, Bo Pang, Chris Quirk for technical advice and the anonymous reviewers for helpful comments.  This paper is based upon work supported in part by the NSF grant IIS-0910664.}
\bibliographystyle{abbrv}

\end{document}